\documentclass[10pt,twocolumn,letterpaper]{article}
\usepackage[utf8]{inputenc}
\usepackage{cvpr}              
\usepackage{graphicx}
\usepackage[dvipsnames]{xcolor}

\definecolor{cvprblue}{rgb}{0.21,0.49,0.74}
\usepackage[pagebackref,breaklinks,colorlinks,citecolor=cvprblue]{hyperref}

\def\paperID{29} 
\def\confName{CVPR}
\def\confYear{2024}

\title{OCC-MLLM:Empowering Multimodal Large Language Model For the Understanding of Occluded Objects}

\author{
Wenmo Qiu\\
University of Toronto\\
wenmo.qiu@mail.utoronto.ca\\
\and
Xinhan Di\\
Giant Network AI Lab\\
dixinhan@ztgame.com
}
\begin{document}
\maketitle
\begin{abstract}
There is a gap in the understanding of occluded objects in existing large-scale visual language multi-modal models. Current state-of-the-art multimodal models fail to provide satisfactory results in describing occluded objects for visual-language multimodal models through universal visual encoders. Another challenge is the limited number of datasets containing image-text pairs with a large number of occluded objects. Therefore, we introduce a novel multimodal model that applies a newly designed visual encoder to understand occluded objects in RGB images. We also introduce a large-scale visual-language pair dataset for training large-scale visual-language multimodal models and understanding occluded objects. We start our experiments comparing with the state-of-the-art models. \end{abstract}
\section{Introduction}
\label{sec:intro}
The latest multimodal dialogue models \cite{chen2023shikra,gong2023multimodal,yang2023mm,chen2023lion,lin2024moe,liu2024visual,alayrac2022flamingo,jin2022expectation,gao2023llama,li2023blip,wu2023visual}, such as MiniGPT-4 \cite{zhu2023minigpt4} and mPLUG-Owl \cite{ye2023mplug} showed that despite significant progress, their description of large-scale language models for occluded objects remains unsatisfactory.

\begin{figure*}[htbp]
    \centering
    \includegraphics[width=\textwidth]{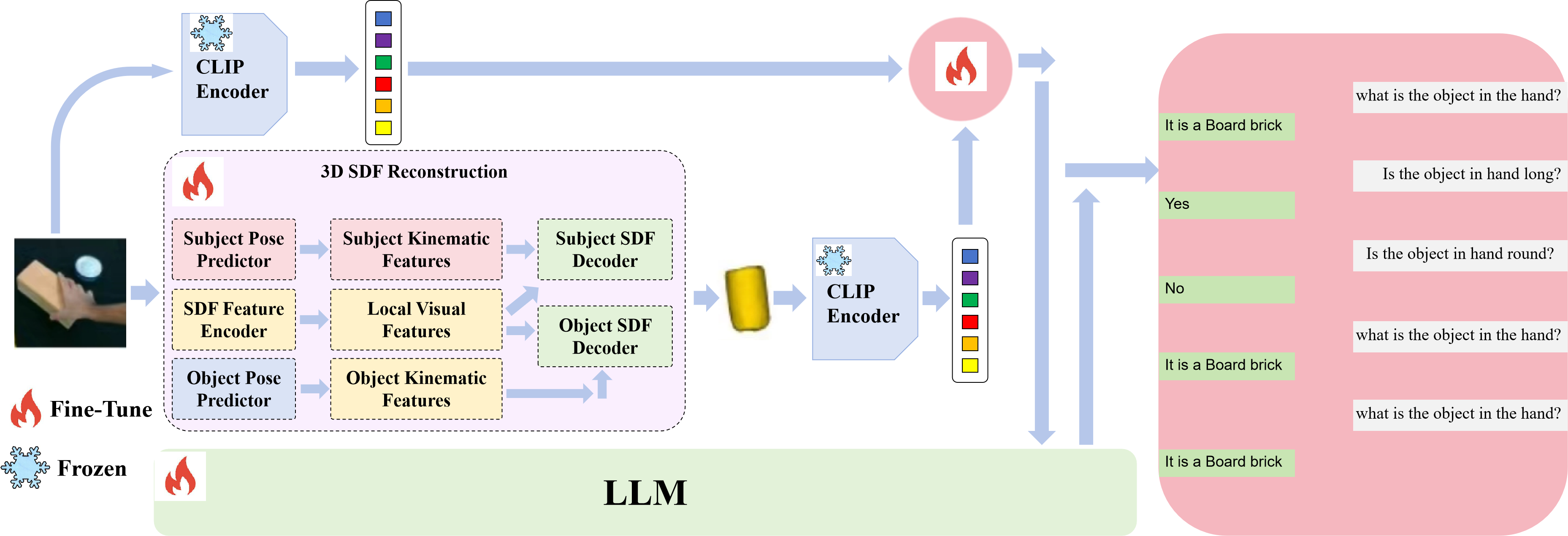}
    \caption{Overview of the Proposed Multi-Modal Vision-Language Model for the Occluded Objects.}
    \label{fig:fig1}
\end{figure*}

Therefore, we propose OCC-MLLM, a visual language model (shown in Figure \ref{fig:fig1}) designed to understand occluded objects in image conversations. To achieve this goal, we developed a visual encoder module consisting of the common CLIP model \cite{radford2021learning} and the proposed 3D model \cite{chen2023gsdf}. Additionally, a dataset of $600,000$ image-text pairs was created and released.


\section{Method}
First, we formulate the generative process of the proposed MLLM, named Occlusion-Aware Multimodal Large Language Model (OCC-MLLM), for occlusion-aware descriptions of objects at hand. Second, we introduce the formulation details of each proposed OCC-MLLM module. Third, the proposed occlusion loss is calculated, and an occlusion-aware training strategy for large multi-modal language models is introduced. We represent the generation process of the proposed OCC-MLLM into three parts: input formula, model forwarding, and decoding.

\subsection{Formulation of OCC-MLLM Generation}
\subsubsection{Input Formulation} 
The input of the proposed OCC-MLLM consists of images and text. Putting aside specific architectural differences, OCC-MLLM generally applies a visual encoder module to extract visual tokens from raw images and uses a cross-modal mapping module to map them to text space as the input of LLM. The mapped visual tokens are used as part of the LLM input along with the text input. The visual tokens are represented as $\mathbf{x}^v=\left\{x_0, x_{1}, \ldots, x_{N-1}\right\}$. $N$ represents the length of the visual token, which is a fixed number in most cases. Similarly, the input text is segmented using a tokenizer and expressed as  $\mathbf{x}^p=\left\{x_N, x_{N+1}, \ldots, x_{M+N-1}\right\}$. The image and text tokens are then concatenated as the final input $\left\{x_i\right\}_{t=0}^{T-1}$ where $T=N+M$.

\subsubsection{Model Forward}
First, OCC-MLLM is trained in an autoregressive manner using causal attention masks, with each token predicting its next token based on the previous token, formally:
\begin{equation}
\begin{aligned}
& \mathbf{h}=\operatorname{F_{MLLM^{Occ}}}\left(\mathbf{x}_i\right) \\
& \mathbf{h}=\left\{h_0, h_1, \ldots, h_{T-1}\right\}
\end{aligned}
\end{equation}
where $\mathbf{h}$ represents the output hidden states of the last layer of the
$\operatorname{F_{MLLM^{Occ}}}$.

Second, the hidden state $h$ is projected by applying the vocabulary head $\mathcal{H}$ via $\operatorname{F_{MLLM^{Occ}}}$. Get the predicted logits (probability) of the next token, and the calculation is as follows:

\begin{equation}
p\left(x_t \mid x_{<t}\right)=\operatorname{SoftMax}\left[\mathcal{H}\left(h_t\right)\right]_{x_t}, \quad x_t \in \mathcal{X},
\end{equation}

where $x_{<t}$ is represented to simplify the sequence $\left\{x_i\right\}_{i=0}^{t-1}$ and $\mathcal{X}$ is represents as the whole vocabulary set.

\subsubsection{Decoding} After applying logits $p\left(x_t \mid x_{<t}\right)$, several decoding strategies have been developed, including greedy decoding, Beam Search \cite{boulanger2013audio}, DoLa, etc. The decoded tokens are concatenated to the last one of the original input text for the next generation round until the end of the generation. The proposed OCC-MLLM applies a beam search strategy \cite{boulanger2013audio} is a decoding strategy based on cumulative scores.

\subsection{Dual Visual Encoder Module}
In forwarding the proposed OCC-MLLM, we designed a new visual encoder module, which consists of two visual encoders. The first visual encoder is the joint CLIP \cite{radford2021learning}, which is used to extract the visual embedding (token) $x_v$ from the RGB input $\mathbf{x}_{\mathrm{v1}}$ without a specific occlusion representation. The second visual encoder is used to provide a representation of the occluded object visual embedding(token) $\mathbf{x}_{\mathrm{v2}}$. Then, the combined representation is calculated as follows:
\begin{equation}
\mathbf{x}^{v}=\alpha \cdot \mathbf{x}^{v1}+(1-\alpha) \cdot \mathbf{x}^{v2}
\label{eq3}
\end{equation}

\begin{figure}
    \centering
    \includegraphics[width=0.35\textwidth]{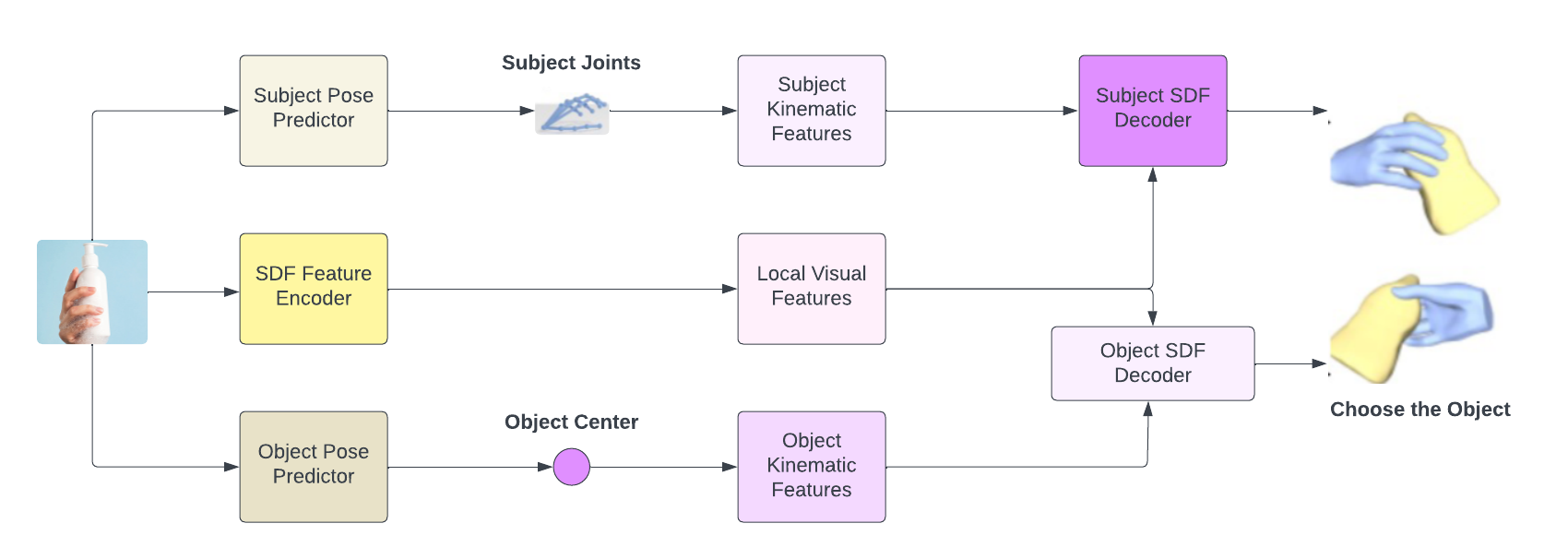}
    \caption{Overview of the proposed second visual encoder reconstruction model
$f_{3D}$. This method reconstructs a mesh of realistic subjects and occluded objects from a single RGB image}
    \label{fig:fig2}
\end{figure}

where $\alpha \in[0,1]$ represents the transparency level of the visual embedding, $\mathbf{x}^v$ represents the merged embedding.

\subsection{Visual Embedding For Occluded Objects}
For the second visual encoder to provide the visual embedding (token) $\mathbf{x}_{\mathrm{v2}}$ of the occluded object, we designed the second visual encoder $f_{3D}$, which is composed as follows:


\begin{figure*}[htbp]
    \centering
    \includegraphics[width=\textwidth]{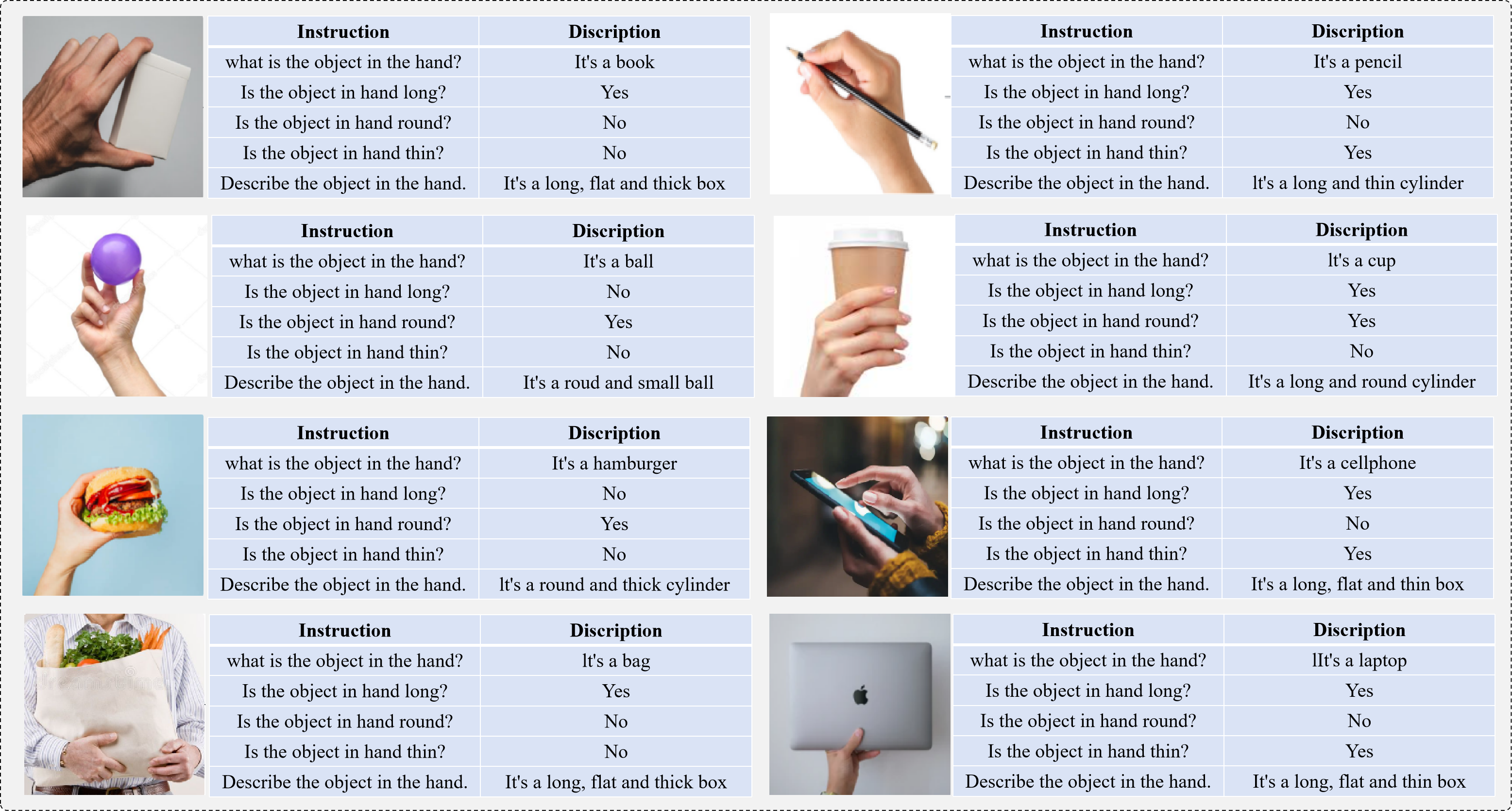}
    \caption{Custom dataset example. The object is occluded. There are five instructions and five corresponding descriptions.}
    \label{fig:fig4}
\end{figure*}

In the first step, a representation of the signed distance function (SDF) \cite{chen2023gsdf} of the occluded object in 3D space is calculated (shown in Figure. \ref{fig:fig2}). This representation is merged into a combination of kinematic and visual features. The SDF of occluded objects and subjects is calculated as follows:
\begin{equation}
\begin{aligned}
& \operatorname{SDF}_{\text {subject}}(v)=f_s\left(\left[e_v ; e_h\right]\right), \\
& \operatorname{SDF}_{\text {object}}(v)=f_o\left(\left[e_v ; e_o\right]\right),
\end{aligned}
\end{equation}
where $f_s$ and $f_o$ are the subject SDF decoder and the object SDF decoder, respectively, $v$ represents the 3D point.

In the second step, we apply the calculated SDFs of bodies and objects for 3D mesh reconstruction (shown in Figure \ref{fig:fig2}). The computed object $\operatorname{SDF}_{\text {object}}(v)$ already contains the visual representation of the object under occlusion. We reconstruct the 3D mesh $M_{obj}$ of the occluded object and then project it into the 2D RGB space $I_{obj}$. Then, to make the 2D visual representation $I_{obj}$ easy to use with large language models, we use the visual embedding of  $\mathbf{x}_{\mathrm{v2}}$ as the extracted embedding of the CLIP model \cite{radford2021learning}. The above calculation is expressed as follows:

\begin{equation}
\begin{aligned}
& M_{obj} = f_{recon}(\operatorname{SDF}_{\text {object}}(v))\\
& I_{obj} = f_{proj}(M_{obj})\\
& \mathbf{x}_{\mathrm{v2}} = f_{CLIP}(I_{obj})\\
\end{aligned}
\end{equation}
\section{Dataset}
We collect a large-scale dataset of occluded objects to train the proposed multimodal large language model to understand them.

\subsection{Dataset Overview}
We released a custom dataset (OCC-HO) containing 600,000 image-text pairs. This dataset was released to describe occluded objects, and to the best of our knowledge, it is for text descriptions of occluded objects. Besides, we manually calculate the occlusions that about a quarter of the objects are occluded on average,   

It is important to note that the annotations of each sample are manually checked. Furthermore, we apply the proposed dataset in the instruction tuning stage. All input images are resized to $224 \times 224$. (Shown in Figure \ref{fig:fig4}). 

\subsection{Dataset Annotation}
We have provided $5$ questions for each image in this dataset. These five questions are: "What's the object in the hand?"; "Is the object in the hand round?"; "Is the object in the hand long?"; "Is the object in the hand thin?"; and "Describe the object in the hand". They are all based on the category, shape, and specific description of the objects in their hands. 

Firstly, we used GPT4V \cite{zhu2023minigpt} to provide preliminary answers to the five questions raised regarding the images. Then, manually check the answers to each image. Manual correction and completion of the answers to the image questions will be done for incorrect or unanswered images. Finally, all the image questions and answers are organized into image pairs to construct a complete dataset of images and texts for occluding objects. 

In addition, we also utilized a 3D reconstruction method \cite{chen2023gsdf} to reconstruct these occluded objects and obtained 2D images containing only objects, further improving our dataset. In this way, the constructed dataset includes images of occluded objects and two image text datasets that only contain images of unobstructed objects after 3D reconstruction.
\section{Experiments and Results}
\subsection{Experiments on GPT4v\cite{openai2023gpt}}
We first test the performance of GPT4v\cite{openai2023gpt} on the testing part of the proposed dataset. Four instructions are applied to test each sample in the testing dataset. And the accuracy is demonstrated in the Table \ref{tab1}. As Table \ref{tab1} shows, the accuracy of the GPT4v\cite{openai2023gpt} is low. In detail, the accuracy for the instruction $1$(What's the object in the hand?) is $0.0361$, the accuracy for the instruction $2$(Is the object in the hand round?) is $0.6705$, the accuracy for the instruction $3$(Is the object in the hand long?) is $0.6290$, the accuracy for the instruction $4$(Is the object in the hand thin?) is $0.5370$. It demonstrates that GPT4V\cite{openai2023gpt} cannot achieve satisfactory results for the occluded objects. 

\subsection{Experiments on MiniGPT4-V2\cite{chen2023minigpt})}
To effectively evaluate the dataset proposed for occlusion object text description, we fine-tuned two epochs for MiniGPT4-V2\cite{chen2023minigpt}. The hyperparameter settings for fine-tuning MiniGPT4-V2\cite{chen2023minigpt} are set as the following: The batch size is $16$; The learning rate is $0.00002$; The weight attenuation coefficient is $0$. In addition, to verify the effectiveness of the constructed occluded dataset. As Table \ref{tab2} shows, in comparison with GPT4V\cite{openai2023gpt}, the accuracy is higher for instruction $1$, the accuracy is about the same for instruction $2$, instruction $3$ and instruction $4$. The visual encoder of the proposed MiniGPT4-V2\cite{chen2023minigpt} is the common clip encoder\cite{radford2021learning}. (Shown in Figure \ref{fig:fig1}). It demonstrates that fine-tuning on a classical multi-modal large language model\cite{lin2024moe} with a single joint clip encoder\cite{radford2021learning} improves the accuracy of the instructions from $0.0361$ to $0.3209$. However, $0.3209$ is still not satisfactory.       


\subsection{Experiments on the Proposed SDF Encoder\cite{chen2023gsdf}} 
Then, we explore the ability of the SDF encoder\cite{chen2023gsdf} for the test description of the occluded objects. At the stage $1$, we pretrain the SDF encoder\cite{chen2023gsdf} for the task of 3D reconstruction\cite{chen2023gsdf} from a single image. At stage $2$, we fine-tune the SDF  encoder\cite{chen2023gsdf}, which loads the weights of the reconstruction\cite{chen2023gsdf} and then fine-tune the encoder for the task of object classification.

In detail, we use each image of the occluded object in the training dataset and the category of the corresponding object for training. In the testing phase, we calculate the accuracy of the occluded objects given a single image of the occluded object. As Table \ref{tab2} demonstrates, the accuracy of the instruction $1$ is further improved from $0.3209$ to $0.5194$. We will continue fine-tuning the proposed SDF encoder\cite{chen2023gsdf} for the tasks corresponding to the instruction $2$-$4$.     

\begin{table}[htbp]
    \centering
    \captionsetup{font={small, bf}} 
    \caption{Experimental results of GPT4V and MiniGPT4-V2 for the proposed dataset}
    \begin{tabular}{ccc} 
        \toprule[\heavyrulewidth] 
        Model & GPT4v(Zero-shot) & MiniGPT4-V2 \\
        \midrule[\lightrulewidth] 
        Instruction 1 & 0.0361 & 0.3209 \\
        Instruction 2 & 0.6705 & 0.6184 \\
        Instruction 3 & 0.6290 & 0.5381 \\
        Instruction 4 & 0.5370 & 0.6017 \\
        \bottomrule[\heavyrulewidth] 
    \end{tabular}
    \label{tab1}
\end{table}

\begin{table}[htbp]
    \centering
    \captionsetup{font={small, bf}} 
    \caption{Experimental results of classification of the object category(Instruction 1) for SDF Encoder}
    \begin{tabular}{ccc} 
        \toprule[\heavyrulewidth] 
        Encoder & Task & Accuracy\\
        \midrule[\lightrulewidth] 
        SDF & Instruction 1 & 0.5194 \\
        \bottomrule[\heavyrulewidth] 
    \end{tabular}
    \label{tab2}
\end{table}

\subsection{Future Experiments}
As the above results demonstrated, the proposed SDF encoder\cite{chen2023gsdf} is promising for understanding the occluded objects. We will explore this encoder's ability in subsequent experiments. 

Firstly, the SDF encoder\cite{chen2023gsdf} continues to be fine-tuned for the task of the instruction $2$, instruction $3$ and instruction $4$. Secondly, the SDF encoder is merged with a classical large language model\cite{lin2024moe} to provide the text description of the occluded objects. Finally, the SDF encoder\cite{chen2023gsdf} and the common clip encoder\cite{radford2021learning} are merged as the equation \ref{eq3} shown, and the proposed dual visual encoder module is applied in a classical multi-modal large language model \cite{lin2024moe} for the description of the occluded objects.  



{
    \small
    \bibliographystyle{ieeenat_fullname}
    \bibliography{main}
}


\end{document}